\documentclass[journal]{IEEEtran}

\usepackage[utf8]{inputenc}
\usepackage[T1]{fontenc}
\usepackage{cite}
\usepackage{amsmath,amssymb,amsfonts,amsthm}
\usepackage{mathtools}
\usepackage{graphicx}
\usepackage{textcomp}
\usepackage{booktabs}
\usepackage{multirow}
\usepackage{array}
\usepackage[dvipsnames,table]{xcolor}
\usepackage{adjustbox}
\usepackage{float}
\usepackage{algorithm}
\usepackage{algpseudocode}
\usepackage[hidelinks]{hyperref}
\algrenewcommand\algorithmicindent{1.0em}
\algrenewcommand\algorithmicrequire{\textbf{Input:}}
\algrenewcommand\algorithmicensure{\textbf{Output:}}
\newcommand{\algsection}[1]{\Statex \smallskip\noindent\textcolor{blue!70!black}{\texttt{//~}\textit{#1}}}

\def\BibTeX{{\rm B\kern-.05em{\sc i\kern-.025em b}\kern-.08em
    T\kern-.1667em\lower.7ex\hbox{E}\kern-.125emX}}

\theoremstyle{definition}

\begin{document}

\title{Macro-Operator Generation and Predicate Selection for TAMP Operator Learning}

\author{Can~Emir~Bora and Emre~Ugur%
\thanks{C.~E.~Bora is with the Department of Computer Engineering, Bogazici University, Istanbul 34342, Turkey (e-mail: can.bora@std.bogazici.edu.tr). E.~Ugur is with the Department of Computer Engineering, Middle East Technical University, Ankara, Turkey, and also with the Department of Computer Engineering, Bogazici University, Istanbul 34342, Turkey (e-mail: emre.ugur@bogazici.edu.tr).}%
\thanks{This work was supported by TUBITAK (The Scientific and Technological Research Council of Turkey) ARDEB 1001 and STAR programs (project number: 120E274), and by the European Union's Horizon Europe Research and Innovation Programme under Grant Agreement No. 101136067 (INVERSE).}%
\thanks{Corresponding author: Can Emir Bora (e-mail: can.bora@std.bogazici.edu.tr).}%
}

\markboth{}{Bora and Ugur: Macro-Operator Generation and Predicate Selection for TAMP Operator Learning}

\maketitle

\begin{abstract}
Creating symbolic operators by hand is one of the main bottlenecks in deploying Task and Motion Planning systems (TAMP). Recent works show that these operators can instead be learned from demonstration data. Existing methods, however, typically learn each action in isolation and cannot capture the recurring multi-step structure of manipulation tasks, so the search becomes intractable on long sequential tasks. A further inefficiency arises in the symbolic state: every provided predicate is evaluated at every search node, even when it never appears in any learned operator. We present a system that addresses both problems. Its central component is the automatic generation of macro-operators, composite actions that compress a recurring sequence of individual actions into a single planning step. Our system discovers causally linked action pairs from the training data, where one action produces exactly the condition that the next one requires, and turns each pair into a new operator. Alongside this, our system prunes every predicate that no learned operator references, which shrinks the symbolic state evaluated at each search node. Together, these changes shorten the effective planning horizon, and the benefit they bring grows with the length of the task. Across four TAMP domains, our method reaches up to a ${\sim}4.6\times$ planning speedup compared to the baseline method, namely Learning Operators for TAMP. More importantly, it solves a long sequential task that the baseline cannot solve. Macro-operator discovery thus not only accelerates planning but, in our long-horizon domain, determines solvability in practice. Our code is available at \href{https://github.com/canemirbora4/macro-tamp}{\nolinkurl{github.com/canemirbora4/macro-tamp}}.
\end{abstract}


\IEEEpeerreviewmaketitle

\section{Introduction}
\label{sec:introduction}

\IEEEPARstart{A}{utonomous} robots in real-world environments face complex planning tasks. High-level reasoning involves deciding which objects to interact with and in what sequence, while low-level tasks require precise geometric decisions, such as determining grasp poses, motion trajectories, and placement locations. This tight coupling between discrete choices and continuous parameters makes robotic planning fundamentally challenging~\cite{garrett2021integrated}.

Task and Motion Planning (TAMP)~\cite{cambon2009hybrid} addresses this challenge through a bilevel approach: symbolic planning operators define an abstract transition model that guides search over action sequences, while a low-level search refines these abstract plans by assigning continuous values (e.g., grasp poses, placement locations) to each step. By using classical AI planning methods, TAMP can leverage heuristics to guide the search and quickly discard infeasible plans, exploring far fewer options than brute-force enumeration~\cite{garrett2018ffrob}.

A significant constraint of TAMP systems is their dependence on manually crafted symbolic operators, which necessitate considerable domain knowledge for accurate specification~\cite{zhu2020hierarchical}. Learning can ease this burden, and it has been applied throughout the TAMP pipeline, from samplers that propose grasp poses and placement locations~\cite{kim2017learning,wang2018active,chitnis2019learning} to guidance models that steer the symbolic search toward promising action sequences~\cite{kim2020learning,driess2020deep}. These methods make the search more efficient, but the operators themselves are still written by hand, so the core specification burden remains.

To address exactly this problem, Learning Operators for Task and Motion Planning (LOFT)~\cite{silverLoftPlaceholder} demonstrates that the operators themselves can be learned directly from transition experience, continuing a long line of action-model learning in classical planning~\cite{arora2018review,pasula2007learning,aineto2018learning}. The learned operators follow the style of the Planning Domain Definition Language (PDDL)~\cite{fox2003pddl}, a standard formalism that describes each action through its preconditions and effects, defined over symbolic facts called \emph{predicates}. They are also probabilistic, since each possible effect is assigned a probability of occurring. At planning time, these operators guide a symbolic search that proposes \emph{plan skeletons}, action sequences whose continuous parameters, such as grasp poses and placement locations, are left open and filled in afterwards by a sampling procedure.

Although this shows that operators can be learned effectively for TAMP, the method comes with several assumptions and limitations. The most significant one is that it treats every individual action in isolation and does not sufficiently address the fact that some actions almost always occur together in a fixed causal order, with one action producing exactly the condition that enables the next. Because of this, the planner must rediscover these recurring two-step patterns from scratch at every search node, which inflates the effective planning horizon and causes the branching factor to explode on long-horizon tasks. A second limitation is that every provided predicate is evaluated at every search node even when it never appears in any learned operator, which wastes state parsing effort and enlarges the symbolic search space.

In this work, we address these limitations one by one with the following contributions:

\begin{itemize}
  \item \textbf{Generating macro-operators:} To address the isolated treatment of individual actions, we automatically generate \emph{macro-operators} from the training data, that is, single composite actions that each replace a short and fixed sequence of individual actions. For this, we identify pairs of actions that are causally linked, where the first produces a condition that the second consumes, and register each pair as a new composite operator. This collapses frequent two-step sub-routines into one planning step. As a result, the effective planning horizon shrinks and the branching factor grows more slowly on long-horizon tasks.
  \item \textbf{Pruning unused predicates:} To address the fact that every provided predicate is evaluated at every search node even when many of them never appear in any learned operator, we propose \emph{Post-Learning Predicate Selection (PPS)}, a post-learning step that automatically removes predicates absent from all learned preconditions and effects. This shrinks the symbolic state the planner must parse and reason over at every node.
\end{itemize}

\begin{figure}[t]
    \centering
    \includegraphics[width=1.0\linewidth]{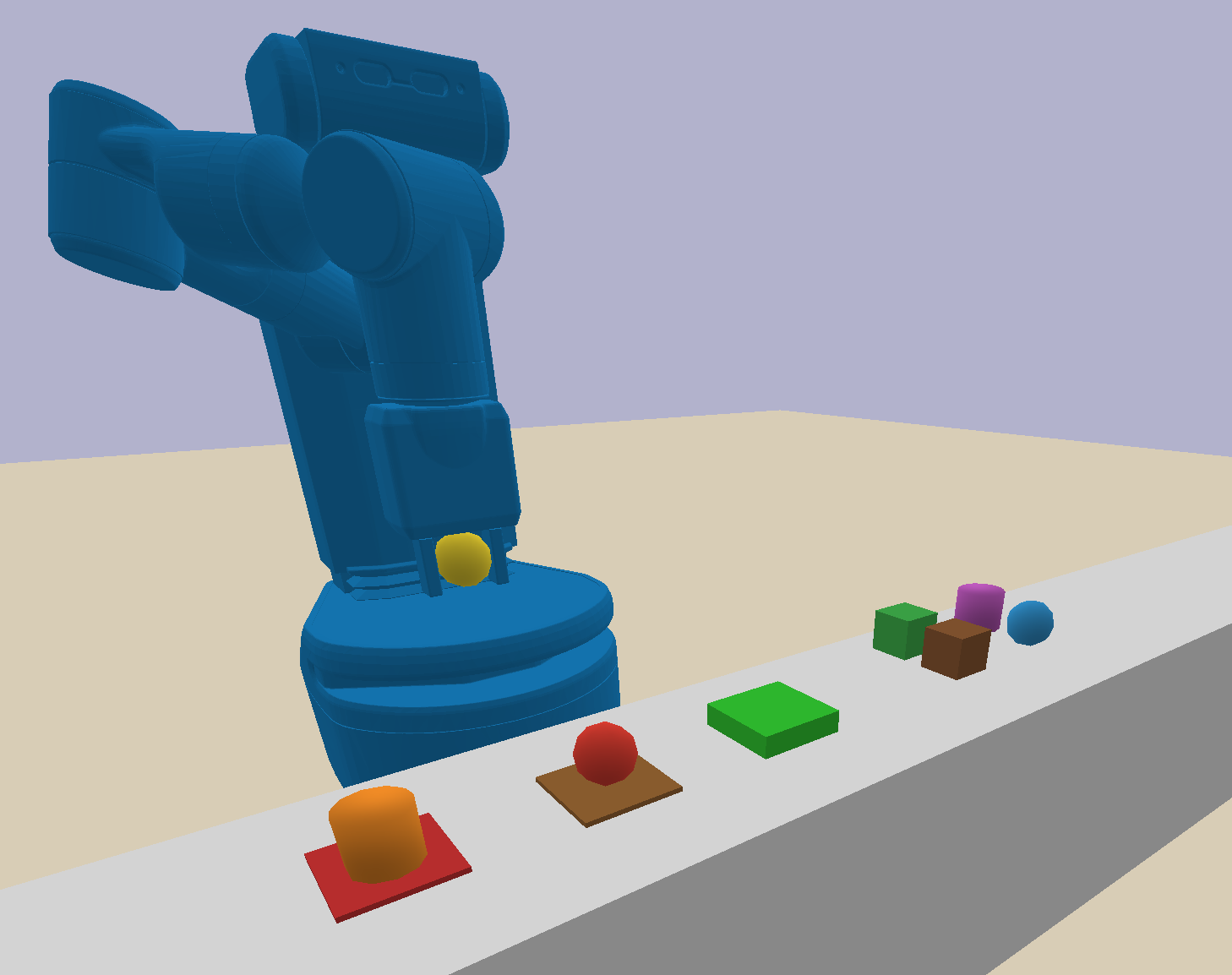}
    \caption{A Kitchen problem in PyBullet. A Fetch robot manipulates seven ingredients, represented by distinct colored shapes, across three workstations: Burner (red pad), Chopping Board (brown pad), and Dish (green pad). Four ingredients wait at the far end of the counter while three are actively being processed. Each ingredient must be picked, chopped, cooked, and plated to satisfy the goal.}
    \label{fig:kitchen_screenshot}
\end{figure}

We evaluate our method on four TAMP domains. Three of them, \textbf{Cover}, \textbf{Blocks}, and \textbf{Painting}, have relatively short plan lengths and are also used in the baseline work~\cite{silverLoftPlaceholder}. In addition, in order to verify the method on long-horizon tasks, we introduce \textbf{Kitchen}, a cooking domain that we designed following the kitchen tasks used in prior TAMP work~\cite{wang2018active,garrett2020pddlstream}, in which six to eight ingredients must each be picked, chopped, cooked, and plated (Figure~\ref{fig:kitchen_screenshot}). Kitchen demands long sequences of interdependent actions, since every ingredient has to pass through several processing stations in a fixed order. Experiments show that our contributions achieve up to ${\sim}4.6\times$ planning speedup and enable solving the Kitchen domain. The baseline cannot solve this domain, and no plan is found there even when the planning time limit is removed. A component-wise ablation study further quantifies the contribution of each proposed component.

\section{Related Work}
\label{sec:related}
\subsection{Task and Motion Planning (TAMP)}
Following the taxonomy of Garrett et al.~\cite{garrett2021integrated}, approaches to TAMP can be broadly categorized based on how symbolic reasoning interacts with continuous variables during the planning process. One family formulates TAMP as a single large-scale optimization problem. It optimizes over hybrid trajectories, that is, trajectories that combine discrete choices with continuous motion, and it integrates logic and geometry directly~\cite{toussaint2015logic,dantam2016incremental}. A second family avoids this joint formulation and instead relies on sampling-based procedures that follow a search-then-sample pattern. These methods first search for a symbolic plan and then sample continuous values to realize it~\cite{kaelbling2011hierarchical,srivastava2014combined,garrett2020pddlstream}. In these methods, symbolic operators, often represented in PDDL~\cite{fox2003pddl}, are used to generate a \emph{plan skeleton}, a sequence of high-level actions with their discrete arguments fixed but continuous parameters left open. A low-level continuous optimizer or sampler then attempts to fill in those continuous arguments. A nearly universal characteristic of these deeply integrated TAMP systems is their reliance on manually authored symbolic operators, whose preconditions and effects must be specified by a human expert~\cite{zhu2020hierarchical}. Our work builds on the premise that these symbolic abstractions can instead be learned automatically from demonstration and interaction data.

\subsection{Learning for TAMP and Operator Discovery}
To reduce the burden of manual specification, learning has been applied to several parts of the TAMP pipeline. At the geometric level, one line of work learns continuous samplers, i.e.\ generative models that propose promising grasp poses or placement locations from the current state, so that the low-level optimizer explores fewer dead ends~\cite{kim2017learning,wang2018active,chitnis2019learning}. At the symbolic level, a complementary line learns search guidance. Kim and Shimanuki~\cite{kim2020learning} learn a value function over relational states that scores symbolic actions and steers the tree search. Driess et al.~\cite{driess2020deep} train a sequence model that predicts whole high-level action sequences from an image of the scene. These approaches learn \emph{how to search} but still rely on hand-written operators describing what each action does.

Closer to our setting is the model-based paradigm of learning the operators themselves, i.e.\ their preconditions and effects. This has a long history in classical planning~\cite{arora2018review}. Examples include noisy deictic rules, a probabilistic rule format that tolerates noisy and uncertain action effects~\cite{pasula2007learning}; LOCM (Learning Object-Centred Models), which extracts action models from plan traces~\cite{cresswell2013acquiring}; STRIPS-style learning, which recovers action models in the classic STRIPS (Stanford Research Institute Problem Solver) representation with the help of classical planners~\cite{aineto2018learning}; and sparse relational transition models learned from data~\cite{xia2019learning}. Whereas these methods assume the symbolic predicates are given, a more recent thread discovers the predicates themselves from robot interaction. Ahmetoglu et al.\ introduce \emph{DeepSym}, which trains a deep encoder-decoder to predict action effects~\cite{ahmetoglu2022deepsym}. This network turns continuous observations into discrete object symbols, and PDDL operators are then extracted over those symbols. Follow-up work extends the idea to multi-object scenes using attention layers that expose object roles and relations, producing relational predicates that generalize across object counts~\cite{ahmetoglu2024relational,ahmetoglu2025symbolic}. Similarly, other efforts learn abstractions tailored to planning, each starting from a different representation: symbols grounded in an agent's skills~\cite{konidaris2018skills}, entity-centric state abstractions for model-based reinforcement learning~\cite{veerapaneni2020entity}, and neuro-symbolic predicates optimized end-to-end for planning performance~\cite{silver2023predicate}. Most relevant to our setting, Huang et al.~\cite{huang2025automated} infer a planning domain, that is, the preconditions and effects of each action, from a small number of demonstrations given at test time. A recent survey places these efforts in the broader context of combining learning and planning~\cite{ugur2025neurosymbolic}. LOFT~\cite{silverLoftPlaceholder} belongs to this operator-learning family but targets TAMP directly: it learns lifted, object-parameterized operators from a handful of demonstrations and uses them inside a sampling-based TAMP planner. However, it suffers from the two limitations set out in Section~\ref{sec:introduction}: each action is learned in isolation, and every provided predicate is evaluated at every search node. We address both in this paper. Since it keeps the operator structure explicit, it is also a natural foundation for the components we propose, and we adopt it as our baseline.

\subsection{Macro-Operators in Planning}
Macro-operators are composite actions formed by chaining several individual actions into a single planning step. They have a long history in classical AI planning: early systems such as MACROPS, a mechanism in the STRIPS planner that stored successful action sequences for later reuse, showed that this can substantially reduce the depth of future searches~\cite{fikes1972learning,korf1985learning}. Subsequent work proposed various criteria for selecting useful macros, for example by analyzing which action pairs recur across problem instances or which state transitions are repeated in training data~\cite{botea2005macro,newton2007learning}.

In robotics, the same idea appears under the name of macro-actions, which group repetitive low-level behaviors into reusable skills and simplify control~\cite{gizzi2019creative,sarathy2020spotter}. Applied to TAMP, macros can collapse multi-step sub-routines such as picking up an object, moving it, and placing it into a single operator, which shortens the effective planning horizon. However, for \emph{learned} probabilistic TAMP operators, the automatic discovery and validation of such macros remain underexplored: existing frameworks learn each action in isolation and ignore the sequential structure present in the demonstration data~\cite{silverLoftPlaceholder,pasula2007learning,aineto2018learning,xia2019learning}. We address this gap by discovering causally linked action pairs directly from training trajectories and registering each pair as a new operator, which the operator learner then treats identically to individual actions during precondition learning and probability estimation.

\section{Method}
\label{sec:method}

\begin{figure}[t]
    \centering
    \includegraphics[width=1.0\linewidth]{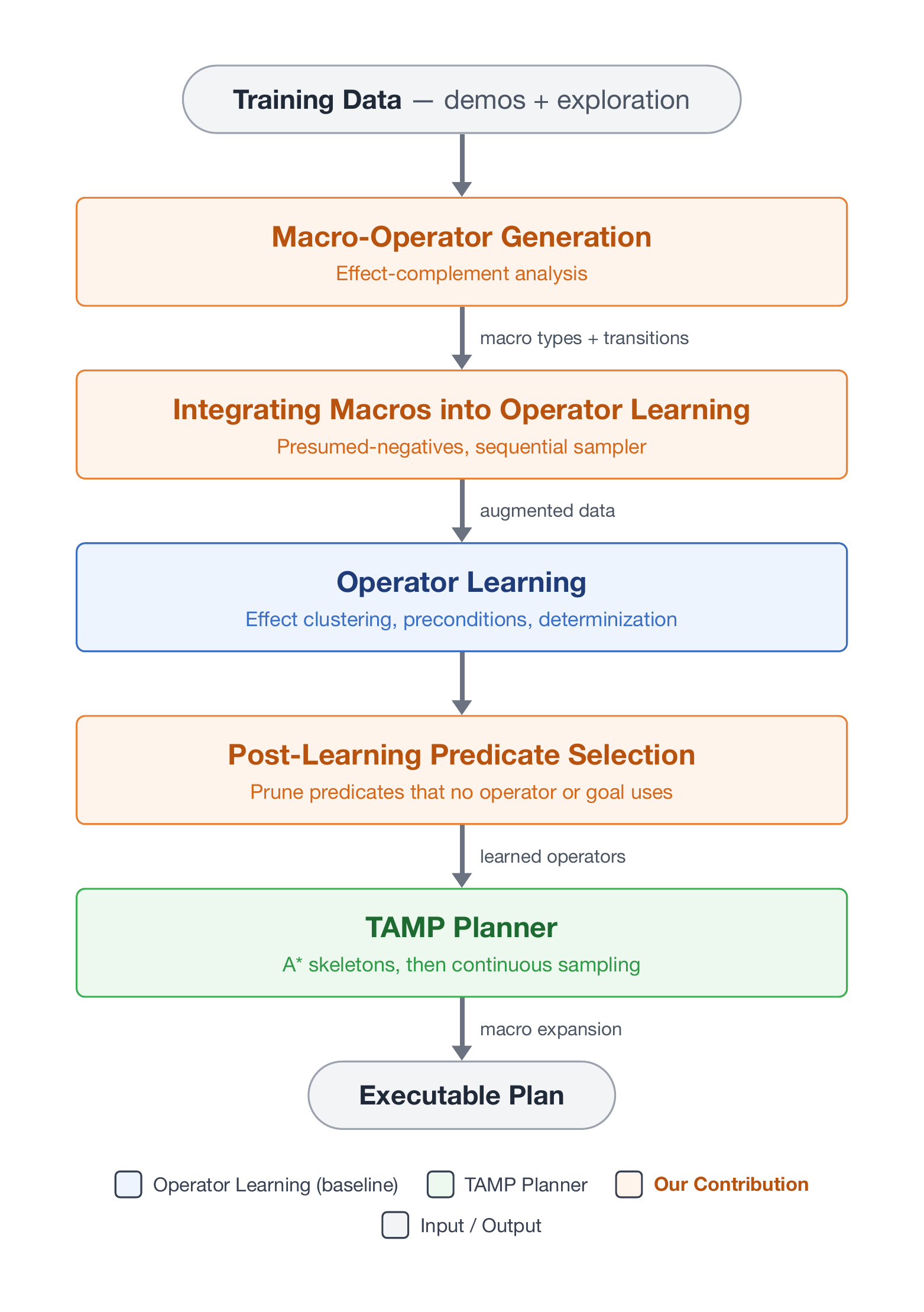}
    \caption{Architecture of the full pipeline, from training data to an executable plan.}
    \label{fig:fullModelPipeline}
\end{figure}

Figure~\ref{fig:fullModelPipeline} shows the complete system we propose, from the training data to an executable plan. Our main contributions are shown with the stages colored with orange boxes, each of which is presented in its own subsection below. Before detailing our contributions, we first give an overview of the background methods used in our pipeline, namely Operator Learning and the TAMP Planner, in the next subsection.

\subsection{Background}
\label{sec:background}

In this section, we provide the background methods on which our work is built. In our TAMP approach, at the high level, a symbolic planner searches with operators written in PDDL. Each operator has preconditions and effects defined over \emph{predicates}, the symbolic facts that abstract the continuous state into a discrete representation. Using A* search~\cite{hart1968formal} with the heuristic~\cite{bonet2001planning} that estimates goal distance by summing the costs of achieving each subgoal independently, this stage returns candidate \emph{plan skeletons}: action sequences whose discrete arguments are fixed but whose continuous parameters are left open. At the low level, a backtracking search turns a skeleton into an executable plan. For each step, it calls domain-specific samplers to propose continuous values, then simulates the action to check feasibility, and reverts to an earlier step whenever the current one fails. Each step is given a fixed number of sampling attempts before it is considered a failure. If no assignment works for a skeleton, the high-level planner produces the next plan.

Next, we describe the \emph{operator learning module}, for which we use LOFT~\cite{silverLoftPlaceholder}. The input is a dataset of low-level transitions, together with a fixed set of user-provided typed predicates. The transitions come from two sources. The oracle demonstrations $\mathcal{D}$ are successful task solutions, whereas the random-exploration set $\mathcal{R}$ is collected by taking random actions from states visited in the demonstrations, thereby providing the negative examples used during learning. The module first converts every continuous state into a symbolic state using a deterministic function, \textsc{Parse}, that returns the subset of predicates that hold in that state. Learning then proceeds in three stages. \emph{(1) Lifted effect clustering} computes the predicates that the action adds and the predicates it deletes. Lifting, here, corresponds to replacing the object instances with typed variables.  Transitions with the same effects but different object instances are grouped into one cluster, and each cluster becomes a candidate operator. \emph{(2) Precondition learning} then finds, for each cluster, the preconditions under which its effect occurs. For this, it scores the sets of observed predicates before each action. A candidate set is scored positively when it is consistently observed in the cluster's own transitions, and negatively when it is observed in transitions for the same action that produced different effects. Note that these negative transitions come mostly from the random exploration data. At the end, the best-scoring set becomes the operator's precondition. \emph{(3) Parameter estimation} computes each effect's probability by counting how often the effect occurs and dividing by the number of times the action was applied: $\hat{p}_k = n_k / N$. Effects whose estimated probability falls below a threshold are discarded. The remaining outcomes are converted into deterministic operators via all-outcome determinization~\cite{yoon2007ff}, yielding one deterministic operator per retained outcome.

\subsection{Our Method}

\subsubsection{\textbf{Macro-Operator Generation}}
\label{sec:macro}

We generate macro-operators automatically through what we call \emph{effect-complement analysis}. The goal is to find pairs of actions that may be causally linked. Two actions form such a pair when the first consistently adds a predicate that the second consistently deletes. The first action thus creates exactly the condition that the second one needs, and then removes it. We detect these pairs directly from the training trajectories as explained in Algorithm~\ref{alg:macro_mine_short}. In the rest of this subsection, we will detail this procedure step by step, with references to this algorithm.

\begin{algorithm}[t]
\small
\raggedright
\caption{Macro-Operator Generation}
\label{alg:macro_mine_short}
\begin{algorithmic}[1]
\Require $\mathcal{D}$: oracle transitions $(x_i, a_i, x_{i+1})$;\quad $\mathcal{R}$: random-exploration transitions $(x_i, a_i, x_{i+1})$;\quad $\mathcal{A}$: action types
\Ensure $\mathcal{A}_{M}$: discovered macro action types;\quad $\mathcal{T}_{+}$: macro transitions

\algsection{Phase 1: Finding consistent effects of each action type}
\For{each transition $(x_i, a_i, x_{i+1}) \in \mathcal{D}$}
  \State $s_i \gets \textsc{Parse}(x_i)$;\quad $s_{i+1} \gets \textsc{Parse}(x_{i+1})$
  \Comment{$\textsc{Parse}$ returns the predicates that hold in a continuous state}
  \State record added predicates $s_{i+1} \setminus s_i$ and deleted predicates $s_i \setminus s_{i+1}$ for action type $a_i$
\EndFor
\For{each action type $a \in \mathcal{A}$}
  \State $\mathrm{Adds}[a] \gets \{\, p \mid \Pr[\,p \text{ added} \mid a\,] > 0.5 \,\}$
  \State $\mathrm{Dels}[a] \gets \{\, p \mid \Pr[\,p \text{ deleted} \mid a\,] > 0.5 \,\}$
\EndFor

\algsection{Phase 2: Causal pairs}
\State $\mathcal{C} \gets \emptyset$
\For{each ordered pair of action types $(a_i, a_{i+1})$ with $a_i \neq a_{i+1}$}
  \State $\mathcal{L}(a_i, a_{i+1}) \gets \mathrm{Adds}[a_i] \cap \mathrm{Dels}[a_{i+1}]$
  \Comment{predicates $a_i$ creates and $a_{i+1}$ removes}
  \If{$\mathcal{L}(a_i, a_{i+1})$ contains a predicate of arity $> 0$}
    \State $\mathcal{C} \gets \mathcal{C} \cup \{(a_i, a_{i+1})\}$
  \EndIf
\EndFor

\algsection{Phase 3: Validation and conflict resolution}
\State $f(a_i, a_{i+1}) \gets$ number of times $(a_i \to a_{i+1})$ occurs consecutively in $\mathcal{D}$ with an atom of arity $>0$ added by $a_i$ and deleted by $a_{i+1}$;\quad $u(a_i, a_{i+1}) \gets$ number of these that leave the state unchanged
\State $\mathcal{V} \gets \{\, (a_i,a_{i+1}) \in \mathcal{C} \mid f > 0,\; u < f \,\}$
\For{each pair with both $(a_i,a_{i+1})$ and $(a_{i+1},a_i)$ in $\mathcal{V}$}
  \State remove from $\mathcal{V}$ the direction with the smaller $f$
\EndFor

\algsection{Phase 4: Build macro operators}
\State $\mathcal{A}_{M} \gets \emptyset$;\quad $\mathcal{T}_{+} \gets \emptyset$
\For{each consecutive $(x_i, a_i, x_{i+1}, a_{i+1}, x_{i+2})$ in $\mathcal{D} \cup \mathcal{R}$ with $(a_i, a_{i+1}) \in \mathcal{V}$}
  \State Create macro action $M$ by unifying the arguments of $a_i$ and $a_{i+1}$
  \State $\mathcal{A}_{M} \gets \mathcal{A}_{M} \cup \{M\}$;\quad $\mathcal{T}_{+} \gets \mathcal{T}_{+} \cup \{(x_i, M, x_{i+2})\}$
\EndFor

\State \Return $\mathcal{A}_{M},\; \mathcal{T}_{+}$
\end{algorithmic}
\end{algorithm}

\textbf{Terminology.} Before describing the procedure, we distinguish two terms used throughout. An \emph{action type} is a template such as \textsc{Pick} or \textsc{Stack}, whose arguments are not yet bound to concrete objects. An \emph{action} is one grounded instance of such a template, such as \textsc{Pick}($b$, $r$), in which the arguments refer to specific objects. The first step is to characterize what each action type consistently changes in the symbolic state.

\textbf{Phase 1: Consistent effects of each action type.} The aim of this phase (lines 1-8) is to find consistent effects of actions from the oracle demonstration set $\mathcal{D}$, as the non-consistent and rare effects are caused by sampling variance rather than by the action itself. We deliberately exclude the random-exploration set $\mathcal{R}$ at this stage, since most of its actions fail and change nothing, which would dilute these ratios and hide effects that the action does produce reliably.
$\mathcal{D}$ includes transitions in the form of $(x_i, a_i, x_{i+1})$: a continuous state $x_i$, the action $a_i$ applied in it, and the resulting continuous state $x_{i+1}$. For every transition, we compute the symbolic states $s_i = \textsc{Parse}(x_i)$ and $s_{i+1} = \textsc{Parse}(x_{i+1})$, where \textsc{Parse} is the function introduced in Section~\ref{sec:background}. A predicate is \emph{added} if it is absent in $s_i$ but present in $s_{i+1}$, that is $s_{i+1} \setminus s_i$. It is \emph{deleted} if it is present in $s_i$ but absent in $s_{i+1}$, that is $s_i \setminus s_{i+1}$. Grouping transitions by their action type, we count how often each predicate is added or deleted. A predicate is considered to be a \emph{consistent add} of action type $a$ if it is added in more than half of $a$'s transitions, and a \emph{consistent delete} if it is deleted in more than half of the times. These two sets, denoted by $\mathrm{Adds}[a]$ and $\mathrm{Dels}[a]$, correspond to the output of Phase~1. The threshold of one half means that an effect must occur more often than not. The validated macros do not depend on its exact value (Section~\ref{sec:discussion}).

\textbf{Phase 2: Causal pairs.} The aim of this phase (lines 9-15) is to find action types that are causally linked, where one action produces a condition that the next one needs. A good macro should join exactly such a pair, since merging them removes an intermediate step the planner would otherwise have to rediscover. 
In order to find the set of predicates that one action consistently creates and another consistently removes, we consider every ordered pair of action types $(a_i, a_{i+1})$, where the subscripts denote the order in which the two would be applied rather than positions in a particular trajectory. For each such pair, we compute the causal-link set $\mathcal{L}(a_i, a_{i+1}) = \mathrm{Adds}[a_i] \cap \mathrm{Dels}[a_{i+1}]$. A non-empty $\mathcal{L}(a_i, a_{i+1})$ means that $a_{i}$ produces a condition that $a_{i+1}$ then consumes, which is exactly the causal link we aim to find. In case a non-empty $\mathcal{L}(a_i, a_{i+1})$ contains only zero-arity predicates, i.e., predicates without object arguments, they are discarded because such global state predicates carry no object-specific causal link, and keeping them would create spurious pairings between unrelated actions. For example, without this rule, \textsc{HandEmpty}() would link every placing action to the next \textsc{Pick}, and a wrong \textsc{Place}$\to$\textsc{Pick} macro would be validated in Painting. The output of this phase is the causal pair set $\mathcal{C}$ that contains action pairs with produced-and-consumed predicates with at least one object argument (e.g. \textsc{Holding}(block) or \textsc{On}(block, table)).

\textbf{Phase 3: Validation and conflict resolution.} The aim of this phase (lines 16-20) is to identify the causal pairs with empirical support, as some pairs in $\mathcal{C}$ might be plausible on paper, but may never actually occur in the data. In order to validate that they are in the data, we count how often each candidate $(a_i, a_{i+1}) \in \mathcal{C}$ appears as two consecutive actions in the oracle demonstrations $\mathcal{D}$. The random exploration set $\mathcal{R}$ is not used here, since it contains no consecutive actions.

An occurrence counts only if the link is real, that is, an atom added by $a_i$ is the same atom that $a_{i+1}$ deletes, with the same objects. An atom is a predicate applied to concrete objects, such as $\textsc{Holding}$($b_1$). A shared predicate name from Phase~2 is not enough, because the same predicate can refer to different objects. As in Phase~2, atoms without arguments, such as $\textsc{HandEmpty}$(), are ignored. Here, we test whether $a_{i+1}$ deletes the atom, not whether it requires it, because the preconditions of $a_{i+1}$ are learned only after this step. Unlike Phase~1, this step counts occurrences rather than ratios, and we denote this count by $f(a_i, a_{i+1})$.

A candidate is validated if it passes two checks. First, it must occur at least once. One occurrence is already meaningful, since it is a real atom link in a successful demonstration. Moreover, a wrong macro would not make a problem unsolvable, because the individual actions stay available to the planner. We analyze this choice and a stricter criterion in Section~\ref{sec:discussion}. Second, the pair must change the symbolic state in at least one occurrence, since a pair that always leaves it unchanged only undoes itself. Writing $u(a_i, a_{i+1})$ for the number of occurrences that leave the state unchanged, this gives the validated set $\mathcal{V} = \{(a_i, a_{i+1}) \in \mathcal{C} \mid f(a_i, a_{i+1}) > 0, u(a_i, a_{i+1}) < f(a_i, a_{i+1})\}$.

After verification, we resolve the conflicting cases when both $(a_i, a_{i+1})$ and $(a_{i+1}, a_i)$ appear in $\mathcal{V}$. Registering both would create two macros for the same underlying pattern, thereby inflating the search branching factor. Whenever both directions are present, we keep only the one with the higher frequency $f$ and discard the other pair. 

\textbf{Phase 4: Building macro operators.} The goal of this phase (lines 21-26) is to turn each verified pair into one usable macro operator.
This requires two components: a single argument list for the macro, and training transitions from which the next stage can learn. Both are obtained by scanning the data for occurrences of the pair.
The first difficulty is that the two sub-actions have separate argument lists. Some objects appear in both actions, others in only one. To behave as a single operator, the macro needs one merged list in which a shared object occupies a single slot. We call this \emph{argument unification}. Continuous parameters, such as a grasp pose or a placement location, are values that each sub-action samples for itself, so they are never merged. Consider a $\textsc{Wash}$ followed by a $\textsc{Dry}$ in the Painting domain: $a_i = \textsc{Wash}(o, w)$ washes object $o$ with the amount of water $w$, and $a_{i+1} = \textsc{Dry}(o, h)$ dries the same object with the amount of heat $h$. Here $o$ is shared, while the continuous parameters $w$ and $h$ belong to one sub-action each, so the merged argument list is $(o, w, h)$. To recall how this list maps back to each sub-action, we store two index vectors: $\mathbf{i}_1 = (1, 2)$ for $\textsc{Wash}$, which uses $o$ and $w$, and $\mathbf{i}_2 = (1, 3)$ for $\textsc{Dry}$, which uses $o$ and $h$. These vectors let the planner recover the original arguments of each sub-action when the macro is later split back into its two steps at execution time.

The macro operator $M$ produced this way is added to the macro action set $\mathcal{A}_M$. We then build the training data for $M$, that is, the transitions from which its preconditions and effects will later be learned. For every consecutive occurrence $(x_i, a_i, x_{i+1}, a_{i+1}, x_{i+2})$ in $\mathcal{D} \cup \mathcal{R}$ with $(a_i, a_{i+1}) \in \mathcal{V}$, we record a positive transition $(x_i, M, x_{i+2})$. This transition skips the intermediate state $x_{i+1}$, so the macro appears to be a single step that takes the world from $x_i$ directly to $x_{i+2}$. These transitions form the set $\mathcal{T}_+$. 

Note that the set of shared objects can differ from one occurrence to another, so the same pair can yield more than one macro, each with its own argument pattern and its own preconditions and effects learned in the next stages.

Generation leaves us with the macro action types $\mathcal{A}_M$ and their positive transitions $\mathcal{T}_+$. At this point a macro is only a typed argument list together with a set of transitions, and not yet a usable operator. It has no learned preconditions or effects, no procedure for proposing continuous parameters for its two sub-actions, and no means of execution on the robot, since it does not belong to the original action set. The next section describes how we integrate the generated macros into the operator learning module to obtain all three.

\subsubsection{\textbf{Integrating Macros into the Operator Learning Module}}
\label{sec:integration}

\textbf{Presumed-Negative Example Generation.} To learn accurate preconditions, the module requires both positive and negative examples: states where the macro's effect occurred, and states where it did not~\cite{pasula2007learning}. Phase~4 already provides the positives $\mathcal{T}_+$, but no negatives exist for macros. For individual actions, such negatives are already available: the random exploration set $\mathcal{R}$ contains many states in which an action was attempted and produced no effect. Macros, however, were not part of the action set when $\mathcal{R}$ was collected, so $\mathcal{R}$ holds no such record for them. We must therefore generate examples for macros ourselves. We call them presumed-negatives, because they are assumed to be negative but this is not verified. We generate them by manufacturing states in which a macro is assumed not to apply.

To this end, for each macro type $M \in \mathcal{A}_M$ we draw a random state $x$ from $\mathcal{R}$ and bind the macro's parameters to objects $\mathbf{o}$ of matching types. We then record the transition $(x, M(\mathbf{o}), x)$, in which the state is left unchanged. These transitions form the presumed-negative set $\mathcal{T}_-$. Since the state and the objects are drawn at random, a sampled state can satisfy the preconditions of the macro. Such a presumed-negative is not a true negative. We do not filter these examples, and we analyze their effect in Section~\ref{sec:discussion}.

\textbf{Learning Macro Operators.} To learn the generated macros without modifying the module, we present each one in the form it already expects for an ordinary action. It takes three inputs: action types $\mathcal{A}$, demonstration transitions $\mathcal{D}$, and random-exploration transitions $\mathcal{R}$ that supply negatives. We augment each with its macro counterpart: $\mathcal{A}' = \mathcal{A} \cup \mathcal{A}_M$ (generated macro types), $\mathcal{D}' = \mathcal{D} \cup \mathcal{T}_+$ (positive transitions from Phase~4), and $\mathcal{R}' = \mathcal{R} \cup \mathcal{T}_-$ (presumed-negatives). Each macro is therefore treated as an ordinary action type, and the module learns its preconditions and effects through the same three stages it applies to individual actions. The only macro-specific addition is a \emph{sequential sampler}, described next.

\textbf{Sequential Continuous Sampling.} A macro must still propose continuous values, such as a grasp pose for its first sub-action and a placement location for its second. These two are not independent: a valid placement depends on where the object was grasped. Sampling them separately would ignore this dependency and produce many infeasible combinations. We therefore give each macro a \emph{sequential sampler} that respects the order of the two sub-actions. It samples the first sub-action's parameters, simulates that sub-action to reach the intermediate state $x_{i+1}$, and then samples the second sub-action's parameters conditioned on $x_{i+1}$. The index vectors from Phase~4 tell the sampler which arguments belong to each sub-action, and each sub-action samples its own continuous parameters. The same index vectors are used once more after planning, when every macro in the returned plan is expanded back into its two sub-actions to give an executable plan over the original action set $\mathcal{A}$.

\subsubsection{\textbf{Post-Learning Predicate Selection (PPS)}}
\label{sec:ips}
The two preceding subsections together form our first contribution. Our second contribution targets the symbolic state itself. As described in the Background, the module turns each continuous state into a symbolic one by evaluating every predicate in the user-provided set $\mathcal{P}$ through \textsc{Parse}. The planner does this at every node it expands, so the cost of $\mathcal{P}$ is paid repetitively throughout the search. The set is also fixed before learning starts, which means the user must choose it without knowing which predicates the learned operators will end up using.

These two facts together create the problem. Once learning has finished, some predicates in $\mathcal{P}$ turn out to appear in no learned precondition and in no learned effect. Such a predicate is still evaluated at every node, and it still enlarges the grounded state, that is, the symbolic state instantiated with all concrete objects in the scene. Yet no operator ever reads it or changes it, so the facts it produces are ones the planner can do nothing with.

PPS removes them once operator learning is complete. Drawing on the principle of feature selection in machine learning~\cite{guyon2003introduction}, we retain only the predicates that occur in at least one learned operator or in the goal of the problem. Writing $\mathcal{O}$ for the learned operator set, $\mathrm{Preds}[o]$ for the predicates appearing in the preconditions or effects of an operator $o$, and $\mathrm{Preds}[G]$ for the predicates appearing in the goal $G$ of the problem being solved, the retained set is
\begin{equation}
  \mathcal{P}' \gets \mathcal{P} \cap \Big( \bigcup_{o \in \mathcal{O}} \mathrm{Preds}[o] \;\cup\; \mathrm{Preds}[G] \Big).
\end{equation}
Goal predicates are always retained. Otherwise, a goal predicate that no learned operator uses would be removed, and the goal could never be reached. Since the goal is known only at planning time, $\mathrm{Preds}[G]$ is added at the start of each planning call. We discuss the effect of this rule on our results in Section~\ref{sec:discussion}. The planner then parses states with $\mathcal{P}'$ in place of $\mathcal{P}$. The operators themselves are left untouched, and the filter costs a single scan over $\mathcal{O}$, which is negligible beside learning.

\section{Experiments}

\subsection{Experimental Setup}
Our evaluation spans four TAMP domains, from simple tabletop
manipulation to complex multi-stage cooking.
\textbf{Cover} is the simplest: place two blocks on targets,
performed in 2--4 steps.
\textbf{Blocks} raises the difficulty by requiring six blocks to be
stacked into goal configurations in approximately 10 steps.
\textbf{Painting} adds object attributes such as color and dryness,
producing 8 operators and plans of approximately 32 steps.
\textbf{Kitchen} is the hardest: six to eight ingredients, each requiring
pick, chop, cook, and plate in strict sequence, yielding plan lengths
that overwhelm search when only individual actions are available.
We designed this domain for this study to evaluate planning over long horizons.

Operator learning is performed with the implementation
of the baseline~\cite{silverLoftPlaceholder}, and PyBullet~\cite{coumans2016pybullet}
serves as the physics backend.
Training data consists of expert demonstrations, where a planner with
full access to the ground-truth operators solves each task, together
with random exploration trajectories that expose the agent to a broader
set of state transitions.
The resulting probabilistic operators learned from this data are then determinized as described
in Section~\ref{sec:background} for use with A* search.
Each domain uses 20 test problems evaluated
over 50 seeds,
with timeouts ranging from 1 to 10\,s per problem.
A seed fixes the test problems and the random number generator of the approach, and all seeds share the same training dataset. Each entry given with a standard deviation is the mean over the 50 seeds, together with the sample standard deviation over the same seeds. For each seed, the success rate is the fraction of its 20 test problems that are solved. The plan time and plan length of a seed are averages over the problems it solves. A problem is unsolved if the planner fails or exceeds the timeout. Unsolved problems are excluded from plan time and plan length and are counted only in the success rate. Plan time, also called planning time, is the wall-clock time of one planning call for a problem. It covers the A* skeleton search, the sampling and simulation of the continuous refinement, and, for our method, the expansion of macros into individual actions. The refinement of our method includes the simulation of the first sub-action by the sequential sampler, so this extra cost is counted in its plan time. The same definition is used for the baseline and for our method. It excludes training, and the timeout limits the same total time. We report the total time rather than the search time alone. The planner generates skeletons one at a time and refines each one before it requests the next, so search and refinement are interleaved.

We compare the original baseline against our complete pipeline,
referred to as \textbf{Ours},
which integrates both contributions presented in
Section~\ref{sec:method}: macro-operator generation
(Sections~\ref{sec:macro} and~\ref{sec:integration}) and Post-Learning
Predicate Selection (Section~\ref{sec:ips}).

\subsection{System Behavior: Learned Operators and Macro Discovery}
\label{sec:qualitative}

We examine macro discovery in detail in the Blocks domain, whose
action set is small enough that every discovered macro can be checked
by hand against what a person would expect.

\textbf{Effect-complement analysis.}
We apply the effect-complement analysis of Section~\ref{sec:macro} to
this domain. Table~\ref{tab:macro_discovery} shows the outcome:
five causal candidates emerge, and requiring a real atom link in the demonstrations leaves two validated macros, \textsc{Pick}$\!\to\!$\textsc{PutOnTable}
(28 occurrences) and \textsc{Pick}$\!\to\!$\textsc{Stack}
(23 occurrences). Both rest on the same causal link: \textsc{Pick} adds
\textsc{Holding}($b$), which \textsc{PutOnTable} and \textsc{Stack}
each delete as their first precondition.
This outcome is the desired one. In this domain, a picked block can
only be placed on the table or stacked on another block, and these are
exactly the two macros the analysis retains. Without any manual
guidance, the procedure thus discovers all meaningful pick-and-place
routines of the domain and discards the other candidates. This result does not depend on the threshold of Phase~1 (Section~\ref{sec:discussion}).

\begin{table}[b]
  \centering
  \small
  \setlength{\tabcolsep}{4pt}
  \caption{Effect-complement macro discovery in the Blocks domain.
    Five causal candidates are identified. The table gives how often each one occurs as two consecutive actions in the demonstrations, and in how many of these occurrences the atom added by the first action is the one deleted by the second. Only candidates with such a link are retained.}
  \label{tab:macro_discovery}
  \begin{adjustbox}{max width=\columnwidth}
  \begin{tabular}{llccc}
    \toprule
    Candidate Pair & Shared Predicate Names (Phase 2) & Occurrences & With Atom Link & Retained \\
    \midrule
    Pick $\to$ PutOnTable & \texttt{Holding}          & $28$ & $28$ & \checkmark \\
    Pick $\to$ Stack      & \texttt{Clear, Holding}   & $23$ & $23$ & \checkmark \\
    PutOnTable $\to$ Pick & \texttt{Clear}            & $19$ & $0$ & $-$ \\
    PutOnTable $\to$ Stack & \texttt{Clear}           & $0$ & $0$ & $-$ \\
    Stack $\to$ Pick      & \texttt{Clear, On}        & $12$ & $0$ & $-$ \\
    \bottomrule
  \end{tabular}
  \end{adjustbox}
\end{table}

\textbf{Unified macro operators.}
Here we examine what the module actually learns for a validated pair. For each
pair, the arguments of the two sub-actions are first unified into a
single list (Section~\ref{sec:macro}), and the module then learns the macro's
preconditions and effects from the augmented dataset, with no manual
guidance. As an example, consider the \textsc{Pick}$\!\to\!$\textsc{Stack}
macro for the case where block $b_1$ starts \textsc{On} another block
$b_3$. Here \textsc{Clear}($b$) means nothing is stacked on block $b$,
\textsc{On}($b$, $b'$) means $b$ rests on $b'$, and \textsc{HandEmpty}
means the gripper holds nothing. The learned
operator takes the following form:

\begin{small}
\begin{quote}
\texttt{MacroPickStack($b_1$:block,\;$b_2$:block,\;$b_3$:block)}\\
\texttt{\quad PRE:\;Clear($b_1$) $\wedge$ Clear($b_2$) $\wedge$ HandEmpty $\wedge$ On($b_1$,$b_3$)}\\
\texttt{\quad ADD:\;On($b_1$,$b_2$) $\wedge$ Clear($b_3$)}\\
\texttt{\quad DEL:\;Clear($b_2$) $\wedge$ On($b_1$,$b_3$)}
\end{quote}
\end{small}

\noindent Read together, the operator says: starting from $b_1$ on
$b_3$ with both $b_1$ and $b_2$ clear and the gripper empty, the macro
ends with $b_1$ on $b_2$ and $b_3$ now clear. This is exactly what a
domain engineer would write by hand for ``pick $b_1$ off $b_3$ and
stack it on $b_2$'', with the intermediate \textsc{Holding} state
absorbed inside the macro. A second variant, for the case where $b_1$
starts \textsc{OnTable} rather than on another block, is discovered
automatically. The precondition search treats the two situations as
distinct patterns without any manual guidance.

\textbf{Planning implications.}
Adding these macros brings the total action set from 4 individual
operators to 7. Executed plan lengths stay roughly the same ($10.5 \pm 1.0$
steps after macro expansion versus $10.9 \pm 1.1$ for the baseline), but
planning is considerably faster, with the plan time dropping from $0.295 \pm 0.239$\,s to
$0.064 \pm 0.034$\,s. Because each macro collapses two decisions into
one, the effective branching factor at every step is reduced, and the
planner reaches goal-relevant states with fewer search nodes.
A representative solved plan illustrates this:

\begin{small}
\begin{quote}
\texttt{[MacroPickPutOnTable($b_5$, pose),}\\
\texttt{\phantom{[}MacroPickPutOnTable($b_4$, pose),}\\
\texttt{\phantom{[}MacroPickStack($b_4$, $b_3$),\;MacroPickStack($b_5$, $b_4$), \ldots]}
\end{quote}
\end{small}

\noindent In the Kitchen domain, the same discovery process
validates 3 causal pairs, namely \textsc{Pick} followed by each of the
three placement actions (\textsc{PlaceOnBoard}, \textsc{PlaceOnBurner},
\textsc{PlaceOnDish}). The operator learner splits each of these macros into two operators, one for each location the ingredient is picked from, which yields 6 macro operators and 16 in
total (10 individual\,+\,6 macro). Without these macros, a problem with six to eight ingredients requires 48 to 64
individual steps, and the expanded plans our pipeline returns average
$55.6$ steps. This value does not vary across seeds, since every seed contains the same mix of problems with six, seven and eight ingredients. This plan length overwhelms the planner
entirely. With macro operators, however, the same goals are achieved
in $0.052 \pm 0.006$\,s across all 20 test problems. To confirm that
the baseline failure is not merely a timeout issue, we ran it with
the timeout disabled; no plan was found after 30 minutes on a single
Kitchen problem.

\subsection{Baseline vs.\ Full Pipeline}
\label{sec:comparison}

In this section we compare our full pipeline against the baseline on
all four domains, in terms of success rate, planning time, and the
structure of the learned operator sets
(Tables~\ref{tab:planning_perf} and~\ref{tab:structural}).

\textbf{Success rates.}
Both approaches achieve $100.0 \pm 0.0$\% on Painting and Cover. In Blocks,
the baseline achieves $99.6 \pm 1.4$\%. The four failures across the 50 seeds are timeouts. In each of them, the planner tries between 179 and 2842 skeletons before the time limit, and all of them fail in refinement. Our full pipeline
achieves a perfect $100.0 \pm 0.0$\% on this domain. It solves the same four problems in 0.1 to 3.2 s and needs 6 skeletons per problem on average, against 28 for the baseline. The most consequential difference arises in
Kitchen: the baseline achieves $0.0$\%
success, while our macro-augmented pipeline solves all 1000 test
instances (20 problems $\times$ 50 seeds) without a single failure.
These numbers show that macros do not cost reliability. On the
three short domains our pipeline is at least as reliable as the
baseline. On Kitchen the gap is not small but total, since the
baseline solves nothing there.

\textbf{Planning time.}
The impact of macro operators is most visible in Blocks, where our
method achieves $0.064 \pm 0.034$\,s versus the baseline's
$0.295 \pm 0.239$\,s, a ${\sim}4.6\times$ speedup in plan time. This speedup does not come from a few slow problems. Our method is faster in 46 of the 50 seeds, and the speedup remains $3.4\times$ when the ten slowest baseline problems are removed. The baseline needs more than one second on 66 of its 996 solved problems. These problems occur in 36 of the 50 seeds. Our method needs more than one second on only 5 problems. Refinement also takes about one third of the plan time in both approaches (31\% for the baseline and 33\% for ours). The speedup is therefore not a shift of cost between search and refinement. In Kitchen, our method plans in $0.052 \pm 0.006$\,s while
the baseline fails entirely. Painting shows a small gain
($0.030 \pm 0.023$\,s versus $0.026 \pm 0.006$\,s), as it is solved
near-instantly regardless of the approach. In Cover, both approaches need $0.001 \pm 0.000\,\text{s}$. Plans there are only two to four steps long, so the planner already finishes in about a millisecond and macros have almost no depth left to remove. Both approaches solve every Cover problem with the first skeleton they try. In all 50 seeds, our method learns the same macro, with \textsc{HandEmpty} as its only precondition and \textsc{Covers}(block, target) as its effect. Across
the four domains the pattern is clear. The benefit of macros grows
with plan length: no gain on two-step tasks, a speedup on ten-step
tasks, and the difference between failure and success on the longest
one.

\begin{table}[t]
  \centering
  \small
  \setlength{\tabcolsep}{3pt}
  \caption{Planning performance of the baseline versus our full
    pipeline. Plan time and plan length are mean\,$\pm$\,std over 50 seeds. For each seed, they are averaged over the problems it solves
    (20 test problems per seed), so unsolved problems are excluded. ``Failed'' indicates $0.0\%$ success
    or no available measurement.}
  \label{tab:planning_perf}
  \begin{adjustbox}{max width=\columnwidth}
  \begin{tabular}{lcccc}
    \toprule
    & \multicolumn{2}{c}{\textbf{Plan Time (s)}} &
      \multicolumn{2}{c}{\textbf{Plan Length (steps)}} \\
    \cmidrule(lr){2-3}\cmidrule(lr){4-5}
    Domain & Baseline & Ours & Baseline & Ours \\
    \midrule
    Cover     & $\mathbf{0.001 \pm 0.000}$ & $\mathbf{0.001 \pm 0.000}$  & $2.6 \pm 0.0$   & $2.6 \pm 0.0$  \\
    Blocks    & $0.295 \pm 0.239$ & $\mathbf{0.064 \pm 0.034}$ & $10.9 \pm 1.1$  & $10.5 \pm 1.0$ \\
    Painting  & $0.030 \pm 0.023$ & $\mathbf{0.026 \pm 0.006}$ & $32.1 \pm 0.4$  & $32.1 \pm 0.4$ \\
    Kitchen   & Failed            & $\mathbf{0.052 \pm 0.006}$ & Failed          & $55.6 \pm 0.0$ \\
    \bottomrule
  \end{tabular}
  \end{adjustbox}
\end{table}

\begin{table}[t]
  \centering
  \small
  \setlength{\tabcolsep}{3pt}
  \caption{Structural properties of learned operator sets: operator
    count, predicate count after PPS, and training time (mean over 50 seeds). Operator
    learning succeeds in every domain, including Kitchen, where it is
    planning rather than learning that fails without macros.}
  \label{tab:structural}
  \begin{adjustbox}{max width=\columnwidth}
  \begin{tabular}{lcccccc}
    \toprule
    & \multicolumn{2}{c}{\textbf{Operators}} &
      \multicolumn{2}{c}{\textbf{Predicates}} &
      \multicolumn{2}{c}{\textbf{Train Time (s)}} \\
    \cmidrule(lr){2-3}\cmidrule(lr){4-5}\cmidrule(lr){6-7}
    Domain & Baseline & Ours & Baseline & Ours & Baseline & Ours \\
    \midrule
    Cover     & 4  & 5  & 5  & \textbf{3}  & 0.013 & 0.020 \\
    Blocks    & 4  & 7  & 6  & \textbf{5}  & 0.052 & 0.148 \\
    Painting  & 8  & 9  & 14 & 14          & 3.735 & 4.124 \\
    Kitchen   & 10 & 16 & 8  & 8           & 0.270 & 0.411 \\
    \bottomrule
  \end{tabular}
  \end{adjustbox}
\end{table}

\textbf{Structural properties.}
Table~\ref{tab:structural} shows operator counts, predicate counts
after PPS, and training times. Macro generation adds 3 operators in
Blocks (one Pick$\to$PutOnTable and two Pick$\to$Stack variants),
1 in Cover, 1 in Painting, and 6 in Kitchen. The number of macro operators can exceed the number of validated pairs. The operator learner splits a macro into several operators when its effects depend on the starting situation, for example on whether the picked block rests on the table or on another block. PPS (Section~\ref{sec:ips})
reduces the active predicate count in Cover ($5\!\to\!3$) and Blocks
($6\!\to\!5$) (\textsc{IsBlock} and \textsc{IsTarget} in Cover, \textsc{IsBlock} in Blocks) by removing predicates
unreferenced by any learned operator. This directly shrinks the
grounded state representation computed at each A* node. In Painting
and Kitchen, all provided predicates
appear in the learned rules, so PPS has no effect. Training times
scale with operator set complexity: Painting requires ${\sim}4$\,s
due to its 8-operator, 14-predicate state space, while all other
domains train in under 0.5\,s. Macro generation adds modest overhead
(e.g., $0.052\!\to\!0.148$\,s in Blocks) for generating macro transitions
and learning preconditions for the additional operators.
The cost of the method is therefore small. At most six operators
are added, and training stays below five seconds in every domain. Both
costs are paid once, before planning starts. The benefit returns at
every node of every later search.

\subsection{Ablation Study: Isolating Component Contributions}
\label{sec:ablation}

We test a \emph{subtractive} ablation: start from the full model and
remove one component at a time. This isolates the contribution of
each component. We run the ablation on Blocks, where both components contribute but the effect is modest, and on Kitchen, where the macro
component is the deciding factor.

\textbf{Blocks domain.}
Removing macros has the largest effect, as Table~\ref{tab:ablation_blocks} shows. The success rate decreases
from $100.0$ to $99.6 \pm 1.4$\% and planning time grows
$4.7\times$, from $0.063$ to $0.299$\,s. Removing PPS leaves the success
rate at $100.0$\% but adds one predicate back to the state, which carries a
small cost at every A* node and raises planning time from $0.063$ to
$0.066$\,s. The two components work at different levels. PPS removes a
fixed cost from every node. Macros reduce how many nodes are visited.

\textbf{Kitchen domain.}
Table~\ref{tab:ablation_kitchen} shows that Kitchen separates the two components more clearly. Removing PPS
leaves performance essentially unchanged, since every provided predicate
already appears in some learned operator and there is nothing left to
prune. Removing the macros, however, takes the success rate to
$0.0 \pm 0.0$\% across every seed and problem, with no plan found even
when the timeout is lifted (Section~\ref{sec:qualitative}). The two
ablations together explain what macro operators provide. Their gain is
not a fixed percentage. It grows with the length of the task, and
in Kitchen it is the only reason a plan is found.

\begin{table}[h]
  \centering
  \small
  \setlength{\tabcolsep}{2pt}
  \caption{Subtractive ablation on Blocks (50 seeds). Removing the
    macro component is the only modification that degrades both
    success rate and planning time. Entries are defined as in Table~\ref{tab:planning_perf}.}
  \label{tab:ablation_blocks}
  \begin{adjustbox}{max width=\columnwidth}
  \begin{tabular}{lcccc}
    \toprule
    Variant & Success (\%) & Plan Time (s) & Operators & Predicates \\
    \midrule
    Ours          & $\mathbf{100.0 \pm 0.0}$ & $\mathbf{0.063 \pm 0.037}$ & 7 & 5 \\
    Ours w/o PPS  & $100.0 \pm 0.0$ & $0.066 \pm 0.037$ & 7 & 6 \\
    Ours w/o Macro & $99.6 \pm 1.4$  & $0.299 \pm 0.242$ & 4 & 5 \\
    \bottomrule
  \end{tabular}
  \end{adjustbox}
\end{table}

\begin{table}[h]
  \centering
  \small
  \setlength{\tabcolsep}{2pt}
  \caption{Subtractive ablation on Kitchen (50 seeds). Only removing
    macro operators causes complete failure. Entries are defined as in Table~\ref{tab:planning_perf}.}
  \label{tab:ablation_kitchen}
  \begin{adjustbox}{max width=\columnwidth}
  \begin{tabular}{lccc}
    \toprule
    Variant & Success (\%) & Plan Time (s) & Operators \\
    \midrule
    Ours          & $\mathbf{100.0 \pm 0.0}$ & $0.052 \pm 0.009$ & 16 \\
    Ours w/o PPS  & $100.0 \pm 0.0$ & $\mathbf{0.050 \pm 0.006}$ & 16 \\
    Ours w/o Macro & $\mathbf{0.0 \pm 0.0}$   & Failed               & 10 \\
    \bottomrule
  \end{tabular}
  \end{adjustbox}
\end{table}

\section{Discussion}
\label{sec:discussion}
In this section, we analyze four design choices of our method. For each one, we report what we measured in our domains.

\textbf{Threshold of Phase~1.} Phase~1 keeps an effect if it occurs in more than half of the transitions of an action type (Section~\ref{sec:macro}). The validated macros do not depend on this value. Any threshold between 0.1 and 0.9 gives the same validated macros in all four domains. The reason is that each pair of action types shows the atom link in either all or none of its occurrences.

\textbf{Minimum count for macro validation.} Phase~3 validates a candidate pair if it occurs at least once (Section~\ref{sec:macro}). In our domains, this simple criterion is sufficient. Every retained pair occurs at least 23 times with an atom link, and every other candidate never does, so any minimum count between 1 and 23 gives the same macros. A higher fixed count would discard valid macros when only a few demonstrations are available. For example, with a quarter of the Blocks demonstrations, a count of one still finds all macros in 99\% of 1000 random subsets, while a minimum count of six loses a valid macro in 58\% of them. For larger or noisier datasets, a minimum share of the occurrences of a pair could be used instead. In our data, both criteria give the same macros. The conflict resolution step of Phase~3 removes no pair in our domains, and we keep it as a safeguard.

\textbf{Effect of presumed-negatives.} Some presumed-negatives of Section~\ref{sec:integration} are not true negatives, since their state satisfies the preconditions of the macro. In our data, this happens for 4\% (Painting) to 62\% (Kitchen) of the presumed-negatives. We tested a stricter version that removes these examples, and the system performed worse. Presumed-negatives can make the learned preconditions more specific, so a macro is used only in situations seen in the demonstrations. For example, \textsc{Pick}$\to$\textsc{PlaceInDish} then also requires \textsc{IsChopped} and \textsc{IsCooked}. In Cover, Blocks and Painting, the learned operators are the same with or without the stricter version. In Kitchen, however, the stricter version lets the macros apply in many more states. The symbolic search then becomes slower and less reliable (1.9\,s instead of 0.05\,s, and 96.0\% instead of 100.0\% success over 10 seeds).

\textbf{Goal predicates in PPS.} PPS always retains the goal predicates (Section~\ref{sec:ips}). In our experiments, however, this rule never changes the result. PPS never removes a goal predicate, because every goal predicate appears in the effect of a learned operator. The rule therefore leaves our results unchanged, but it keeps PPS safe in other domains.

\section{Conclusion}

Learning symbolic operators from data removes one of the main
bottlenecks in deploying TAMP systems, yet existing methods leave
two practical barriers in place: a symbolic state cluttered with
predicates that no learned operator ever uses, and a planning horizon
that search with individual actions alone cannot overcome. This work
addresses both gaps. We proposed an automated macro-operator generation
pipeline that discovers causally linked action pairs through
effect-complement analysis. We also proposed Post-Learning Predicate
Selection, which prunes up to 40\% of the unreferenced predicates and
cuts the parsing overhead paid at every A* node.

Taken together, the experiments suggest that the benefit of macro-operators is not a fixed speedup but a shortened effective planning horizon, whose value scales with the length of the task. In very short, two-step
problems, there is almost no depth left to remove, so our method does not bring any advantage over the baseline method. In longer, for example 10-step problems, our method can achieve a ${\sim}4.6\times$ speedup over the baseline method. 
Moreover, in the Kitchen tasks of 48 to 64 steps, it determines whether a plan is found at all: baseline individual-action planner can not
solve that domain within thirty minutes, while our pipeline solves it
in ${\sim}0.05$\,s. In the tested setting, finding which
actions belong together is therefore a requirement for planning to
work rather than a late optimization. Testing our method on further long-horizon domains is a natural next step to show how generally this holds.

Our current approach leaves several directions open. Our
generation pipeline currently considers only pairs of consecutive
actions. Extending it to longer chains or nested sequences (e.g.,
discovering full pick-wash-dry-place routines) would enable deeper
plan abstraction. A second direction concerns the predicates
themselves: our pipeline prunes unused predicates but cannot create
new ones. Discovering new predicates automatically from sensor data
would reduce the remaining domain engineering effort. Similarly,
the continuous samplers that propose grasp poses and placement
locations are currently hand-designed; learning them from
interaction data would remove another manual component. All
experiments in this work run in simulation using
PyBullet~\cite{coumans2016pybullet}. Deploying the learned operators
on a real robot would introduce noisy perception, imprecise
execution, and sim-to-real transfer challenges that the current
pipeline does not address. Evaluating robustness under these
conditions is an important next step. Finally, the current system
learns offline: it first collects all training data, then mines
macros and learns operators in a single batch. An online variant
that updates its operator set incrementally as the robot encounters
new situations would be more practical for long-lived deployment,
where the task distribution may shift over time.

\bibliographystyle{IEEEtran}
\bibliography{refs}

\end{document}